\documentclass{article}

\usepackage[preprint]{corl_2026}

\usepackage{amsmath,amsfonts,bm}

\def\Figref#1{Figure~\ref{#1}}

\def\Secref#1{Section~\ref{#1}}
\def\eqref#1{equation~\ref{#1}}
\def\Eqref#1{Equation~\ref{#1}}
\def\1{\bm{1}}

\def\rmC{{\mathbf{C}}}

\def\vzero{{\bm{0}}}

\def\vtheta{{\bm{\theta}}}

\def\vc{{\bm{c}}}

\def\vp{{\bm{p}}}

\def\vt{{\bm{t}}}

\def\vv{{\bm{v}}}

\def\vx{{\bm{x}}}

\def\vz{{\bm{z}}}

\def\mI{{\bm{I}}}

\DeclareMathAlphabet{\mathsfit}{\encodingdefault}{\sfdefault}{m}{sl}
\SetMathAlphabet{\mathsfit}{bold}{\encodingdefault}{\sfdefault}{bx}{n}

\newcommand{\pdata}{p_{\rm{data}}}
\newcommand{\E}{\mathbb{E}}
\newcommand{\Ls}{\mathcal{L}}
\newcommand{\R}{\mathbb{R}}

\usepackage{graphicx}
\usepackage{subcaption}
\usepackage{wrapfig}
\usepackage{url}
\usepackage{booktabs}
\usepackage{multirow}

\title{FlowDrive: moderated flow matching with data balancing for trajectory planning}

\author{
  Lingguang Wang \\
  Karlsruhe Institute of Technology \\
  \texttt{lingguang.wang@kit.edu} \\
  \And
  \"{O}mer \c{S}ahin Ta\c{s} \\
  FZI Research Center for Information Technology \\
  \texttt{tas@fzi.de} \\
  \And
  Marlon Steiner \\
  Karlsruhe Institute of Technology \\
  \texttt{marlon.steiner@kit.edu} \\
  \And
  Nick Le Large \\
  Karlsruhe Institute of Technology \\
  \texttt{nick.lelarge@kit.edu} \\
  \And
  Christoph Stiller \\
  Karlsruhe Institute of Technology \\
  \texttt{stiller@kit.edu} \\
}

\begin{document}

\maketitle

\begin{abstract}
Learning-based planners are sensitive to the long-tailed distribution of driving data. Common maneuvers
dominate datasets, while dangerous or rare scenarios are sparse. This imbalance can bias
models toward the frequent cases and degrade performance on critical scenarios. To tackle this problem,
we compare balancing strategies for sampling training data and find reweighting by
trajectory pattern an effective approach. We then present \mbox{FlowDrive}, a flow-matching trajectory planner that
learns a conditional rectified flow to map noise directly to trajectory distributions with few flow-matching steps.
We further introduce moderated, in-the-loop guidance that injects small perturbation between flow steps to
systematically increase trajectory diversity while remaining scene-consistent. On nuPlan and the interaction-focused
interPlan benchmarks, FlowDrive achieves state-of-the-art results among learning-based planners and approaches methods
with rule-based refinements. After adding moderated guidance and light
post-processing (\mbox{FlowDrive*}), it achieves overall state-of-the-art performance across nearly all benchmark splits. Our code is available at \url{https://github.com/einsteinguang/flow_drive_planner}.
\end{abstract}


\section{Introduction}\label{sec:introduction}

Trajectory planning in autonomous driving requires both safety and efficiency.
Traditional planners rely on rule-based methods such as model predictive control and graph- or sampling-based planning. They are interpretable
and safety-driven but often struggle in complex real-world conditions \citep{schwarting2018planning}. Learning-based planners instead learn policies from data
that capture nuanced human driving behaviors \citep{10164673, cheng2024pluto, cheng2023rethinking}, and can rival classical systems on large-scale benchmarks.

Despite progress, real driving data remains long-tailed: common behaviors such as lane-following dominate, while rare but safety-critical cases are underrepresented \citep{karnchanachari2024nuplan}.
This biases planners toward frequent patterns and weakens generalization, especially in dynamic unseen traffic scenarios \citep{9009463}.

Generative models provide a promising solution. Diffusion models \citep{song2019score,ho2020ddpm} generate trajectories
via iterative denoising and can model multi-modal behaviors, but often require many steps or careful guidance for feasibility.
Flow matching \citep{lipman2023flowmatching,liu2023rectifiedflow} offers an alternative generative paradigm.
Instead of iterative denoising, it trains a continuous transformation mapping a simple prior directly to the data
distribution \citep{lipman2023flowmatching}. This enables fast few-step sampling while preserving the ability to model
multi-modal behaviors.

Motivated by the need for diverse yet fast trajectory generation, we propose \emph{FlowDrive}, a flow-matching
planner for autonomous driving. Unlike diffusion planners that produce trajectories via many denoising steps, FlowDrive
learns a continuous motion flow that directly transforms random initial noise into diverse driving trajectories,
yielding faster sampling. However, naively training such a planner on standard driving data can bias it toward common scenarios and neglect underrepresented but critical
cases. Because flow matching learns global transport geometry, it can also be sensitive to data imbalance. We therefore analyze how training-set imbalance affects planning and introduce
a data-balancing method that increases coverage of rare behaviors. We also introduce a mechanism to steer
FlowDrive's output trajectories to diversify the generated candidates. Finally,
we evaluate FlowDrive on the nuPlan \citep{karnchanachari2024nuplan} and interPlan benchmarks
\citep{Hallgarten2024interPlan}.

In summary, the contributions of this paper are:
\begin{itemize}
    \item We identify the impact of unbalanced training data on planning performance and propose a data-balancing
    strategy to improve robustness to rare scenarios. We present \textbf{FlowDrive}, a flow-matching
    trajectory planner that efficiently generates feasible driving trajectories for autonomous vehicles.
    \item We introduce a guidance mechanism that steers FlowDrive's outputs to produce more diverse trajectory samples.
    This enables state-of-the-art performance on the nuPlan and interPlan benchmarks, outperforming prior rule-based,
    learning-based and hybrid baselines.
\end{itemize}

\section{Related work}\label{sec:related-work}

\subsection{Learning-based planning methods}
Imitation learning directly trains models to mimic expert driving. \citet{cheng2023rethinking} show
that careful architecture design and augmentation improve closed-loop performance. \citet{cheng2024pluto} propose
PLUTO, which adds contrastive learning and data augmentation, surpassing rule-based planners on nuPlan. These works show that imitation learning can be competitive, but such planners often
struggle in rare or unseen scenarios.

Reinforcement Learning (RL) offers an alternative. \citet{zhang2025carplanner} introduce CarPlanner, a
consistent autoregressive RL planner that stabilizes training and surpasses imitation methods. \citet{tang2025planr1}
propose Plan-R1, which combines pre-training on expert data with RL fine-tuning using rule-based rewards,
achieving strong results. However, RL methods often rely on carefully designed reward functions and substantial training \citep{hafner2023mastering}.

\subsection{Data balancing for learning-based planners}
Recent works have begun to address the problem of data imbalance. \citet{wang2025safefusion} propose SafeFusion,
which clusters trajectory "vocabularies" and applies balanced sampling with adaptive loss weighting to mitigate
the skew between collision and non-collision cases. Similarly, \citet{parekh2025balancedbc} tackle imbalance in
behavior cloning, showing that conventional weighting biases policies toward frequent behaviors while neglecting
rare but critical ones, and propose meta-gradient reweighting to improve policy robustness. We introduce a cluster-based trajectory reweighting strategy for imitation-learning-based planning.

\subsection{Diffusion and flow-based models for planning}
Diffusion models are widely applied for trajectory generation. \citet{zheng2025diffusionplanner}
combine prediction and planning in a transformer-based diffusion framework. \citet{liao2024diffusiondrive}
introduce DiffusionDrive, a diffusion planner that generates multimodal trajectories with truncated Gaussian priors.
\citet{yang2024diffusiones} propose Diffusion-ES, combining diffusion with evolutionary search to optimize
trajectories under arbitrary rewards. These works highlight diffusion's flexibility, but also its sampling overhead and reliance on external guidance.

Flow matching avoids iterative denoising with many steps \citep{lipman2023flowmatching} and was originally introduced for
realistic image generation. In planning, \citet{xing2025goalflow} propose GoalFlow, a goal-conditioned flow
matching planner that is integrated into an end-to-end autonomous driving pipeline and focuses on goal-oriented guidance.
\citet{nguyen2025flowmp} extend flow matching to second-order planning, yielding smoother motions in robotic manipulation tasks.
However, flow matching remains underexplored for trajectory planning.

\subsection{Guidance for generative models}
Guidance is crucial for controlling generative models. In vision, classifier-based and classifier-free guidance steer
diffusion sampling toward desired attributes with higher fidelity \citep{dhariwal2021diffusion,ho2022classifierfree}.
In driving, \citet{zheng2025diffusionplanner} use a trajectory scoring model to guide diffusion toward safe plans.
\citet{yang2024diffusiones} apply black-box reward guidance for high-reward trajectories. \citet{xing2025goalflow} adapt guidance for
flow matching, conditioning trajectories on goal points. Inspired by these works, we introduce a simple diversity-guidance method for flow-matching planning.

\section{Methodology}\label{sec:methodology}
We formulate trajectory planning as conditional generative modeling. Given a scene context \(\vc\) encoding
the High-Definition map, real-time traffic light information, static objects and dynamic agents, we aim to draw a feasible
future trajectory \(\vx\) for
the ego vehicle. We represent the trajectory as a vector in \(\R^{D}\) (later instantiated as \(H\) waypoints,
flattened to a vector). FlowDrive learns a time-dependent velocity field \(\vv_{\vtheta}(t, \vx, \vc)\) that
transports a simple base distribution \(p_0\) (e.g., Gaussian over trajectory dimensions) to the conditional data
distribution \(p_{\rm data}(\vx\mid\vc)\) by integrating an ordinary differential equation (ODE). We build on
flow matching and rectified flow \citep{lipman2023flowmatching,liu2023rectifiedflow}, which enable straight probability paths and fast few-step sampling.

FlowDrive contains three orthogonal components: (i) a conditional flow-matching planner trained with a rectified path,
yielding feasible trajectories; (ii) a data balancing scheme that mitigates long-tail biases in driving
datasets; and (iii) an inference-time moderated guidance mechanism that increases diversity.

\subsection{Preliminaries: rectified flow and flow matching}\label{subsec:preliminaries}
Let \(\vx\in\R^{D}\) denote a data sample (a trajectory) and \(\vc\) the conditioning context. Let \(\vz\sim p_0\) be
a base sample, with \(p_0=\mathcal{N}(\vzero,\mI)\). Rectified flow (RF) defines a linear,
``rectified'' path between \(\vz\) and \(\vx\):

\begin{equation}
\vx_t \;=\; (1 - t)\,\vz \, + \, t\,\vx, \qquad t\in[0,1].
\label{eq:rf-path}
\end{equation}

This path induces a family of intermediate distributions \(p_t(\cdot\mid\vc)\) connecting \(p_0\) to \(p_{\rm data}(\cdot\mid\vc)\),
governed by the continuity equation with some velocity field \(\vv^*(t,\cdot,\vc)\):

\begin{equation}
\partial_t p_t(\vx\mid\vc) \, + \, \nabla_{\vx}\!\cdot\!\big( p_t(\vx\mid\vc)\, \vv^*(t,\vx,\vc) \big) \,=\, 0.
\label{eq:continuity}
\end{equation}

For the rectified path in \Eqref{eq:rf-path}, the target instantaneous velocity along each pair \((\vz,\vx)\) is constant and equals

\begin{equation}
\vv_t(\vx_t,\vc) \;=\; \vx - \vz, \qquad \text{with } \vx_t \text{ as in \Eqref{eq:rf-path}}.
\label{eq:rf-target}
\end{equation}

Flow matching (FM) \citep{lipman2023flowmatching} trains a parametric vector field \(\vv_{\vtheta}(t,\vx,\vc)\) to match
the target velocity that makes \Eqref{eq:continuity} hold along the chosen path. With the rectified path, the standard
conditional RF/FM objective samples \(t\sim \mathcal{U}[0,1]\), \(\vz\sim p_0\), forms \(\vx_t\) via \Eqref{eq:rf-path},
and minimizes

\begin{equation}
\Ls_{\rm RF}(\vtheta) \,=\, \E_{\substack{(\vx,\vc)\sim \pdata(\vx,\vc),\\ \vz\sim p_0,\; t\sim \mathcal{U}[0,1]}} \\
\Big[\, w(t)\, \big\|\, \vv_{\vtheta}(t, \vx_t, \vc) - (\vx - \vz) \,\big\|_2^2 \Big],
\label{eq:rf-loss}
\end{equation}

where \(w(t)\) is an optional time-weighting schedule. At inference, samples are generated by integrating the learned probability flow ODE
\begin{equation}
	\frac{\mathrm{d}}{\mathrm{d}t}\, \vx_t \,=\, \vv_{\vtheta}(t, \vx_t, \vc), \qquad \vx_0\!\sim p_0, \quad t\in[0,1],
\label{eq:prob-flow-ode}
\end{equation}
typically with a small number of solver steps thanks to the low curvature of rectified paths
\citep{liu2023rectifiedflow,lee2023curvature}. Rectified flow and flow matching are closely related to diffusion models via the probability flow ODE perspective \citep{song2021score}, but avoid simulating stochastic dynamics during training.

In FlowDrive, \(\vx\) represents the ego trajectory, \(\vc\) the scene context, and \(\vv_{\vtheta}\) is parameterized by an encoder--decoder.

\subsection{FlowDrive}\label{subsec:flowdrive}
\begin{figure}[t]
    \centering
    \includegraphics[width=1.0\linewidth]{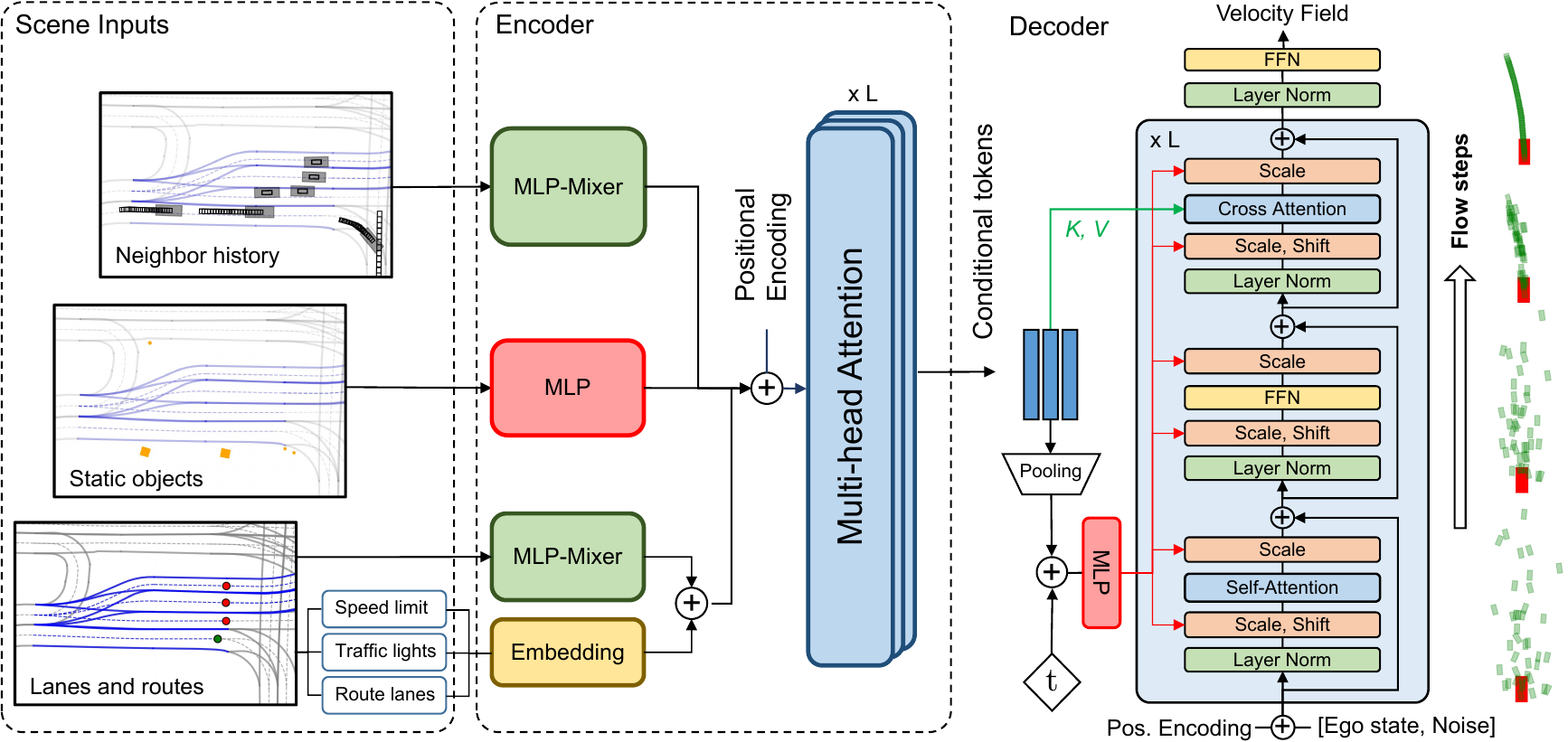}
    \caption{FlowDrive architecture. Left: scene inputs (neighbor history, static objects,
    lanes/routes with traffic lights and speed limits). Middle: encoder with MLP-Mixer branches and multi-head attention fusion.
    Right: DiT-based decoder with adaptive layer norm conditioning and cross-attention to context, predicting the velocity field across flow steps.}
    \label{fig:overview}
\end{figure}

\textbf{Overview.} FlowDrive implements the conditional velocity field \(\vv_{\vtheta}(t,\vx,\vc)\) in \Secref{subsec:preliminaries}
with an encoder--decoder architecture (\Figref{fig:overview}). The encoder aggregates heterogeneous scene inputs \(\vc\)
into a set of context tokens; the decoder predicts the velocity field over the trajectory sequence,
conditioned on time and the encoder tokens. During training, we minimize the rectified flow loss in \Eqref{eq:rf-loss}.
At inference, we integrate the probability flow ODE in \Eqref{eq:prob-flow-ode} with a small number of flow steps.

\textbf{Scene Inputs and Representation.} We follow a scene input representation similar to Diffusion Planner \citep{zheng2025diffusionplanner}.
All coordinates are in the local ego frame, with origin at the current ego position and heading aligned with the ego heading. For a batch of size \(B\): neighbor history has shape \([B, P_n, T_p, F_n]\) with positions,
kinematics, and a one-hot agent type; static objects are \([B, P_s, F_s]\); lanes are polylines \([B, P_\ell, V, F_\ell]\)
with local geometry, traffic-light signals, route membership and per-lane speed limit features. We denote the full context as \(\vc\), and the trajectory
sample as a sequence of \(H\) steps with action dimension \(A\) (\(A=4\) for \(x,y,\cos,\sin\)); we flatten \(\vx\in\R^{H\times A}\)
to \(\R^{D}\) when referring to \Eqref{eq:rf-loss}. Here, \(P_n\) denotes the number of neighbor agents; \(T_p\) the length of
the neighbor history; \(F_n\) the neighbor feature dimension; \(P_s\) the number of static objects; \(F_s\) the static-object feature
dimension; \(P_\ell\) the number of lane segments; \(V\) the number of points per lane segment; \(F_\ell\) the per-point lane feature
dimension.

All scene inputs are normalized to zero mean and unit variance. During training, we randomly perturb ego position, heading, and velocity, following \citet{zheng2025diffusionplanner}.

\paragraph{Encoder.} The encoder builds token embeddings for three branches and fuses them:
\begin{itemize}
    \item \emph{Neighbors.} We use an MLP-Mixer branch \citep{tolstikhin2021mlpmixer,zhang2024premixermlpbasedpretrainingenhanced,zheng2025diffusionplanner} that applies token- and channel-mixing MLPs
    over the temporal dimension and feature channels of each neighbor, followed by average pooling to obtain one token
    per neighbor. A small type embedding is added.
    \item \emph{Static objects.} A projection MLP maps per-object features to hidden tokens. Empty or invalid objects are masked out.
    \item \emph{Lanes and routes.} Another MLP-Mixer branch encodes lane polylines, where per-point geometry is first projected,
    then mixed across points and channels. The token is enriched with: (i) traffic light embedding;
    (ii) speed limit embedding; and (iii) a binary embedding for whether the current lane is part of the global route. We discuss
    the fusion design in Appendix~\ref{subsec:ablation-studies}.
\end{itemize}

The MLP and embedding layers output \(N=P_n+P_s+P_\ell\) tokens of hidden size $d$.
A positional embedding is added to each token, and $L$ multi-head attention blocks with residual MLPs
\citep{vaswani2017attention} fuse them into context tokens \(\rmC\in\R^{N\times d}\) and a binary mask for padded or invalid tokens.

\paragraph{DiT Decoder.} The decoder uses and extends the Diffusion Transformer (DiT) \citep{peebles2023dit}.
Given a noised trajectory (or pure noise) \(\vx_t\in\R^{H\times A}\) at time \(t\), we embed per-step actions with an MLP, add a learned
positional encoding over horizon steps, resulting in $H$ trajectory tokens.
We additionally embed the ego state with an MLP to the same embedding space and append the ego state embedding
in front of the trajectory tokens. We discuss the reason in \Secref{subsec:design-choices}. Each DiT
block uses adaptive LayerNorm-zero (adaLN-Zero) modulation driven by \(\vt\) and the mean pooling of valid context tokens, then
applies: (i) self-attention and an MLP on the trajectory tokens; and (ii) cross-attention to the encoder tokens (keys/values),
considering the token mask. The decoder produces \(\vv_{\vtheta}(t,\vx_t,\vc)\in\R^{(H+1)\times A}\), where the first output
token corresponds to the ego state and is ignored during loss computation w.r.t. the target velocity in \Eqref{eq:rf-target}.

\subsection{Data balancing}\label{subsec:data-balancing}
Although we train on nuPlan, the balancing strategies below are generic. Prior works report severe skew in driving behavior frequencies (e.g., large amounts of
stationary or lane-following samples versus very few rare maneuvers) \citep{zheng2025diffusionplanner,karnchanachari2024nuplan,tas2025wordsmotionextractinginterpretable}.
On \(~10^6\) sampled nuPlan training scenarios, one might observe only \(\sim 1\) sample for \texttt{changing\_lane\_with\_lead}, but \(\sim 350{,}000\) for a simple \texttt{stationary} scenario.
Such imbalance biases a planner toward the frequent modes, undermining robustness in safety-critical cases.

We adopt two complementary sampling strategies.

\begin{figure}[t]
    \centering
    \includegraphics[width=0.99\linewidth]{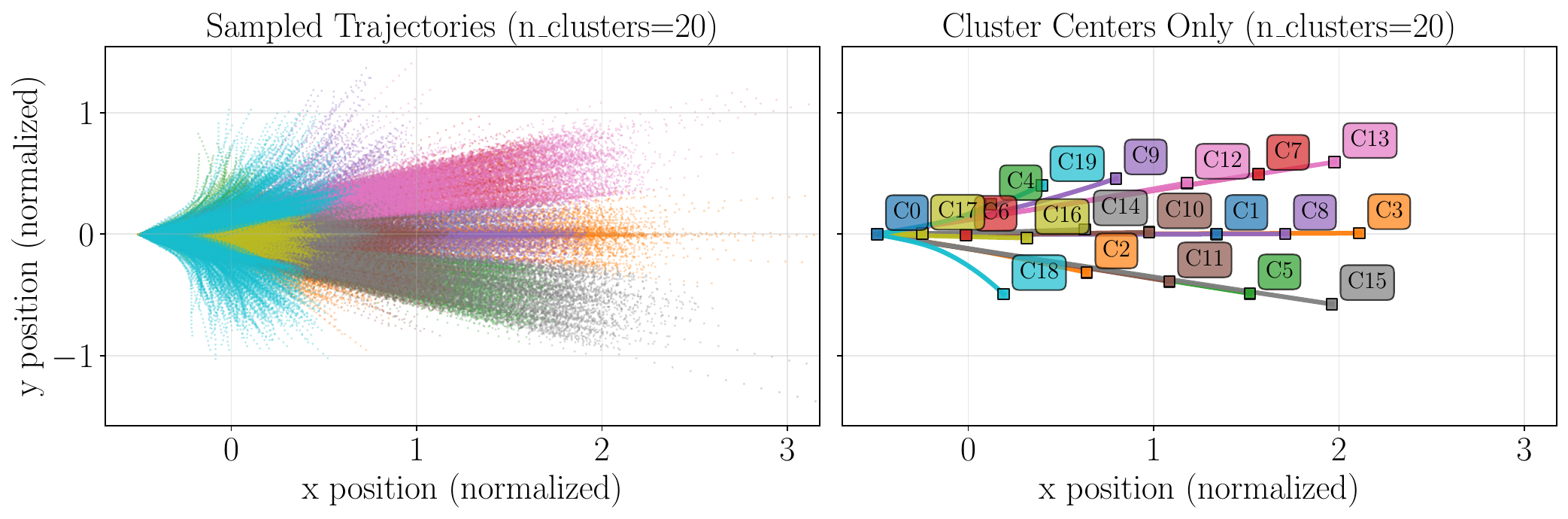}
    \caption{Precomputed clusters of normalized ego futures used for cluster-based sampling. Colored polylines indicate
    cluster centers; translucent points show a subset of sampled trajectories per cluster.}
    \label{fig:clusters}
\end{figure}

\paragraph{Scenario-based Sampling.} Each training sample is assigned a scenario type (for nuPlan: merging, turning,
stopping for traffic lights, etc.). We count scenario frequencies and use inverse-frequency weights so underrepresented
types are sampled more often. Concretely, if \(f_s\) is the fraction of samples belonging to scenario type \(s\),
we define a weight \(w_s = 1/(f_s + \varepsilon)\) and normalize \(\{w_s\}\) to mean 1.0. During training, the
dataloader draws indices proportional to these weights.

\paragraph{Cluster-based Sampling.} To directly balance the distribution of ground-truth ego trajectories, we precompute clusters of
normalized trajectories over the horizon and upweight samples from rare clusters. Specifically,
we embed each ego trajectory into a fixed-length vector (stacked \(x,y\) across time), run k-means to obtain \(K=20\) clusters,
and store the cluster centers and per-cluster statistics (mean and standard deviation). Ablation study of the number of clusters on
the final performance is provided in Table~\ref{tab:clusters-ablation}. \Figref{fig:clusters} visualizes
the precomputed cluster centers and sampled trajectories. Given the cluster assignment for each training
sample, we compute inverse-frequency weights per cluster, analogous to the scenario-based scheme, and use them in the
weighted sampler. This encourages exposure to a wider variety of motion patterns during training.

Cluster-based sampling substantially flattens the per-cluster distribution and increases coverage of rare motion patterns
(see \Figref{fig:cluster-sampling-comparison} in Appendix~\ref{subsec:sampling-distribution} for a detailed comparison).
An ablation comparing these strategies by closed-loop score is presented in Appendix~\ref{subsec:ablation-studies}.

\subsection{Moderated guidance on flow matching trajectories}\label{subsec:moderated-guidance}
While FlowDrive achieves strong closed-loop results, single-pass trajectories can lack lateral diversity in scenes where lateral maneuvers would improve the score.
Prior state-of-the-art planners on nuPlan—e.g. the Diffusion Planner
\citep{zheng2025diffusionplanner} and the PDM planner \citep{chitta2023pdm}—therefore apply post-hoc,
rule-based lateral offsets on the final trajectories.
In contrast, we introduce a \emph{moderated guidance} that injects small, structured perturbations \emph{inside} the
flow integration.

\paragraph{Moderated Guidance.}

Let the (noised) trajectory state at flow time \(t\) be
\[
\vx_t = \big(\vp_{t,1},\dots,\vp_{t,H}\big), \qquad
\vp_{t,h} = (x_{t,h},\, y_{t,h},\, \cos\theta_{t,h},\, \sin\theta_{t,h}) \in \R^{4},
\]
so \(\vx_t \in \R^{H\times 4}\), where \(h=1,\dots,H\) indexes the trajectory horizon steps.
We operate only on the positional part
\(\mathbf{p}_{t,h}^{\text{pos}}=(x_{t,h},y_{t,h})\). The orientation components \((\cos\theta_{t,h},\sin\theta_{t,h})\) are left unchanged.
Define unit tangent and normal (left-hand) directions w.r.t. the current heading angle of the ego vehicle $\theta$ (one
can use the average heading of the lane centerline as well):
\[
\hat{\tau}=(\cos\theta,\, \sin\theta), \qquad
\hat{n}=(-\sin\theta,\, \cos\theta).
\]

Let \( \mathcal{T}_g \subset [0,1] \) (discrete flow times) be a set
of flow times at which guidance is injected. The binary schedule $\alpha(t)=\mathbf{1}\{t\in \mathcal{T}_g\}$
activates guidance only at those steps. A monotonically increasing horizon weight
\(\beta(h) = h/H\) means larger perturbation is injected into later waypoints.

\begin{figure}[t]
    \centering
    \begin{minipage}[c]{0.42\textwidth}
        \centering
        \includegraphics[width=\linewidth]{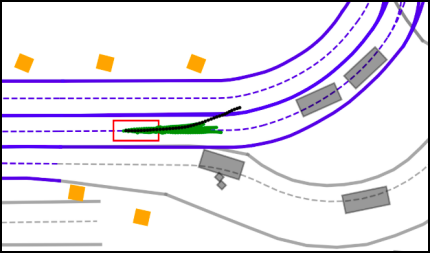}
    \end{minipage}\hfill
    \begin{minipage}[c]{0.54\textwidth}
        \caption{Beyond lane-following green trajectories, moderated lateral offsets induce multi-modality (black overtaking trajectory).}
        \label{fig:overtake}
    \end{minipage}
\end{figure}

Given lateral and longitudinal magnitudes \(\delta_{\text{lat}}\) and \(\delta_{\text{lon}}\),
the moderated update applied before evaluating the next velocity field in the ODE solver (\Eqref{eq:prob-flow-ode}) is
\begin{equation}
\label{eq:moderated-update}
\mathbf{p}'^{\text{pos}}_{t,h}
= \mathbf{p}^{\text{pos}}_{t,h}
+ \alpha(t)\,\beta(h) \left(
\delta_{\text{lat}}\, \hat{n} +
\delta_{\text{lon}}\, \hat{\tau}
\right), \qquad h=1,\dots,H.
\end{equation}

which updates $\vx_t$ to $\vx'_t$. This in-the-loop perturbation lets the learned velocity field reconcile the guided displacement with scene context, unlike post-hoc shifts.
Lateral offsets (\(\delta_{\text{lat}}\)) promote modal diversity (overtaking, nudging), while longitudinal offsets (\(\delta_{\text{lon}}\)) can reshape speed profiles. Longitudinal guidance brought no improvement in our experiments;
thus in all experiments we set \(\delta_{\text{lon}}=0\) and sample \(\delta_{\text{lat}}\in[-1,1]\) (e.g. \([-0.5,-0.25,0,0.25,0.5]\)),
producing the diverse candidates shown in \Figref{fig:moderated-steps} (Appendix~\ref{subsec:injection-step}).

\paragraph{Diversity and Multi-modality.} Lateral offsets often unlock multi-modal behaviors that are otherwise rare in
training data. For instance, in car-following with a slow leading vehicle, small offsets encourage overtake-like
solutions if the adjacent lane is free; see \Figref{fig:overtake}.

We inject guidance at a single flow step (\(|\mathcal{T}_g|=1\)).
Empirically, \(\mathcal{T}_g=\{\tfrac{1}{2}\}\) offers the best trade-off between diversity and feasibility
and is used by default (see Appendix~\ref{subsec:injection-step} for a comparison across injection steps). Unlike manual post-processing, our guidance keeps the model ``in the loop'', letting scene conditioning absorb and correct the perturbation. More examples of the jointly moderated guidance with lateral and longitudinal offsets can be found in \Figref{fig:guidance}.

\section{Experiments}\label{sec:experiments}

\subsection{Benchmarks and experimental setup}\label{subsec:dataset-setup}
We evaluate FlowDrive on the large-scale nuPlan benchmark \citep{karnchanachari2024nuplan} and the interaction-focused interPlan
benchmark \citep{Hallgarten2024interPlan}. On nuPlan, we report results on the official Val14, Test14, and Test14-hard splits
in both non-reactive and reactive modes. InterPlan is a closed-loop driving benchmark based on the nuPlan framework. It modifies original nuPlan scenarios by augmenting agent counts and behaviors so that more intelligent
maneuvers become necessary, such as sudden intruding pedestrians and nudging around parked vehicles.
Both benchmarks use a composite score between 0 and 100 combining safety, progress, and comfort.
All training details and hyperparameters are listed in \Secref{subsec:parameters-flowdrive}.

We evaluate three variants of our planner. \textbf{FlowDrive$^-$}: a purely learning-based variant that executes the
trajectory produced by the decoder directly, without weighted sampling or any post-processing. \textbf{FlowDrive}:
FlowDrive$^-$ with cluster-based sampling.
\textbf{FlowDrive*}: a hybrid variant that applies our moderated guidance during flow integration
(\Secref{subsec:moderated-guidance}) and a light post-processing pass on top of FlowDrive. The post-processing includes:
(i) generating 30 candidates from lateral guidance offsets.
(ii) scoring them with the rule-based scorer from PDM planner \citep{chitta2023pdm},
and executing the top-scoring plan.
We provide an ablation study of the number of candidates on driving scores in Appendix~\ref{subsec:ablation-studies}.

\subsection{Baselines}\label{subsec:baselines}
We compare against representative rule-based, learning-based, and hybrid planners; FlowDrive* is a hybrid variant.
IDM: the classical Intelligent Driver Model for longitudinal car-following and collision avoidance \citep{treiber2000idm}.
PDM-Planner: a modular policy decomposition planner with hand-crafted rules and model-based pathing,
provided as a strong rule-based nuPlan baseline \citep{chitta2023pdm}.
GameFormer: a transformer-based planner that models multi-agent interactions with game-theoretic
inductive bias for safe decision making \citep{gameformer2023}.
PlanTF: a transformer planner that formulates planning as sequence modeling with end-to-end training
on expert demonstrations \citep{cheng2023rethinking}.
PLUTO: an imitation-learning planner enhanced with contrastive learning and augmentations that
achieves strong closed-loop scores on nuPlan \citep{cheng2024pluto}.
Diffusion Planner: a transformer-based diffusion generative planner that denoises trajectories under
flexible guidance \citep{zheng2025diffusionplanner}. Note that we also have a retrained version of Diffusion Planner with our
cluster-based sampling, denoted as ``Diffusion Planner (balanced)'' in \Secref{subsec:ablation-studies}.

\subsection{Quantitative results}\label{subsec:main-results}

Table~\ref{tab:main} presents the quantitative results. Without data balancing, FlowDrive$^-$ still outperforms
previous learning-based planners in almost all columns. With cluster-based sampling,
FlowDrive surpasses them by noticeable margins. Despite using no post-processing, its scores are very close to the strong hybrid baselines
PDM-Hybrid and PLUTO. Adding moderated guidance and post-processing (FlowDrive*) then pushes
performance further to the top, achieving state-of-the-art across nearly all columns
among all planner categories. Note that the scores are mostly extracted from previous publications. On interPlan, we compute the
scores for each baseline using the official code if available.

\paragraph{Generality of Cluster-based Data Balancing.} Our proposed cluster-based data balancing is a
\emph{plug-and-play} for learning-based planners. Retraining the Diffusion Planner with cluster-based sampling (``Diffusion Planner (balanced)'' in Table~\ref{tab:main})
improves its average score across all benchmark splits by \textbf{4.49\%}. Applying the same strategy to FlowDrive$^-$ yields an average
improvement of \textbf{4.94\%} (FlowDrive). Because the procedure operates at the dataloader level and reweights trajectories by ego motion pattern, it is also potentially applicable to end-to-end approaches.

\newcommand{\best}[1]{\textbf{#1}}
\newcommand{\second}[1]{\underline{#1}}
\begin{table*}[t]
    \centering
    \caption{Closed-loop scores on nuPlan Val14, Test14-hard, and Test14 (Non-Reactive, NR; Reactive, R)
    and interPlan. Higher is better. Within each category block, best is \best{bold}; second-best is \second{underlined}.
    A dash ``--'' indicates the result is unavailable or not reproducible due to missing code.}
    \vspace{4pt}
    \resizebox{\textwidth}{!}{
    \begin{tabular}{llcccccc|c}
        \toprule
        \textbf{Type} & \textbf{Planner} & \multicolumn{2}{c}{\textbf{Val14}} & \multicolumn{2}{c}{\textbf{Test14-hard}} & \multicolumn{2}{c}{\textbf{Test14}} & \textbf{InterPlan} \\
                      &                  & NR & R                             & NR & R                                   & NR & R                              & \\
        \midrule
        Expert & Log-replay & 93.53 & 80.32 & 85.96 & 68.80 & 94.03 & 75.86 & 14.76 \\
        \midrule
    \multirow{8}{*}{Learning-based} & PDM-Open & 53.53 & 54.24 & 33.51 & 33.53 & 52.81 & 57.23 & 25.02 \\
     & GameFormer w/o refine. & 13.32 & 8.69 & 7.08 & 6.69 & 11.36 & 9.31 & -- \\
     & PlanTF& 84.27 & 76.95 & 69.70 & 61.61 & 85.62 & 79.58 & 30.53 \\
     & PLUTO w/o refine. & 88.89 & 78.11 & 70.03 & 59.74 & 89.90 & 78.62 & -- \\
     & Diffusion Planner & 89.87 & 82.38 & 75.99 & 69.22 & 89.19 & 82.93 & 24.71 \\
     & FlowDrive$^-$ (ours) & 90.30 & 81.91 & 77.07 & 69.46 & 90.19 & 83.47 & \second{31.42} \\
    & Diffusion Planner (balanced) & \second{90.57} & \second{84.86} & \second{77.34} & \second{71.12} & \second{90.35} & \second{85.73} & 29.26 \\
     & \textbf{FlowDrive} (ours) & \best{91.21} & \best{85.37} & \best{77.86} & \best{73.09} & \best{91.02} & \best{87.28} & \best{36.96} \\
    \midrule
    \multirow{7}{*}{\shortstack{Rule-based \\ \& Hybrid}} & IDM & 75.60 & 77.33 & 56.15 & 62.26 & 70.39 & 74.42 & 31.20 \\
     & PDM-Closed & 92.84 & 92.12 & 65.08 & 75.19 & 90.05 & 91.63 & 41.23 \\
     & PDM-Hybrid & 92.77 & 92.11 & 65.99 & 76.07 & 90.10 & 91.28 & 41.61 \\
     & GameFormer & 79.94 & 79.78 & 68.70 & 67.05 & 83.88 & 82.05 & 11.07 \\
     & PLUTO & 92.88 & 89.88 & \second{80.08} & 76.88 & 92.23 & 90.29 & \second{42.87} \\
     & Diffusion Planner w/ refine. & \second{94.26} & \second{92.90} & 78.87 & \best{82.00} & \second{94.80} & \second{91.75} & -- \\
& \textbf{FlowDrive*} (ours) & \best{94.81} & \best{92.96} & \best{81.86} & \second{81.96} & \best{95.02} & \best{93.96} & \best{44.05} \\
     \bottomrule
    \end{tabular}}
    \label{tab:main}
\end{table*}

Full ablation studies covering data balancing strategies, flow training/inference steps, and post-processing components
are presented in Appendix~\ref{subsec:ablation-studies}. More qualitative results are presented in \Secref{subsec:qualitative-challenging}.

\section{Conclusion}\label{sec:conclusion-discussions}

We introduced FlowDrive, a conditional flow-matching planner that generates trajectories efficiently via
a learned rectified flow. Two components enable its strong performance: (i) cluster-based
data balancing that reweights training samples by ego-trajectory pattern, increasing the coverage of rare
motion modes; and (ii) moderated, in-the-loop guidance that systematically expands diversity while keeping the
trajectories scene-consistent. On nuPlan and interPlan, FlowDrive outperforms all previous learning-based methods by large margins.
With moderated guidance and light post-processing (FlowDrive*), it achieves state-of-the-art on nearly all benchmark splits while remaining efficient.

\section{Limitations}\label{sec:limitations}

Despite the strong benchmark results, FlowDrive still has several limitations. First, the raw trajectories from
FlowDrive can be jerky, especially for stronger lateral maneuvers (see the first row of \Figref{fig:compare-lane-change}). Future work should investigate smoothness constraints or second-order action formulations \citep{nguyen2025flowmp}. Second, our data-balancing method clusters trajectories only in future ego-trajectory \([x,y]\) space. However, motion patterns can also be characterized by lane topology, traffic-rule context, or interaction intent.
Extending cluster-based balancing with map-relative features could better target such corner
cases. Finally, current moderated guidance injects fixed-magnitude offsets between flow steps.
Future work could combine richer safety- and intent-aware objectives \citep{feng2025guidanceflowmatching},
interpretable control vectors \citep{tas2025wordsmotionextractinginterpretable}, and RL fine-tuning methods for flow
models \citep{mcallister2025flowmatchingpolicygradients,zhang2025reinflow} to shift the learned velocity field toward more robust and comfortable behaviors.

\acknowledgments{
    The authors gratefully acknowledge the computing time provided on the high-performance computer HoreKa
    by the National High-Performance Computing Center at KIT (NHR@KIT).
    This center is jointly supported by the Federal Ministry of Education and Research and the Ministry of Science,
    Research and the Arts of Baden-Württemberg,
    as part of the National High-Performance Computing (NHR) joint funding program
    (\url{https://www.nhr-verein.de/en/our-partners}).
    HoreKa is partly funded by the German Research Foundation (DFG).
}

\bibliography{flow_drive_arxiv}

\appendix
\section{Appendix}\label{sec:appendix}

\subsection{Training details and parameters}\label{subsec:parameters-flowdrive}
We train FlowDrive on 1M sampled scenarios (same as Diffusion Planner and PLUTO) from nuPlan dataset. The training is done on 8
NVIDIA RTX 6000 Ada GPUs. The training takes about 40 hours for 400 epochs. The encoder and decoder of the model are
wrapped in an exponential moving average (EMA) model \citep{moralesbrotons2024exponentialmovingaverageweights}. We find that a simple discrete Euler solver suffices to generate trajectories with good quality.
Table~\ref{tab:training-parameters} provides detailed hyperparameters and configuration used for FlowDrive.

\begin{table}[ht]
    \centering
    \caption{Training parameters and hyperparameters for FlowDrive.}
    \label{tab:training-parameters}
    \footnotesize
    \begin{tabular}{llc}
        \toprule
        \textbf{Type} & \textbf{Parameter} & \textbf{Value} \\
        \midrule
        \multirow{3}{*}{Data Preprocessing} & Num. dynamic objects & 32 \\
         & Num. static objects & 5 \\
         & Num. lane segments & 70 \\
        \midrule
        \multirow{8}{*}{Training} & Effective batch size & 2800 \\
         & Effective learning rate & 5e-4 \\
         & Learning rate scheduler & cosine \\
         & Optimizer & AdamW \\
         & Weight decay & 1e-5 \\
         & Warmup epochs & 3 \\
         & EMA power & 0.99 \\
         & ODE solver & Euler \\
        \midrule
        \multirow{2}{*}{Inference} & Flow training steps & 2000 \\
         & Flow inference steps & 8 \\
        \midrule
        \multirow{4}{*}{Encoder} & Num. Attention layers & 3 \\
         & Hidden dim & 192 \\
         & Num. heads & 6 \\
         & Dropout & 0.1 \\
        \midrule
        \multirow{5}{*}{Decoder} & Num. DiT layers & 3 \\
         & Hidden dim & 192 \\
         & Num. heads & 6 \\
         & Dropout & 0.1 \\
         & Num. predicted poses & 40 \\
        \bottomrule
    \end{tabular}
\end{table}

\subsection{Effect on Sampling Distribution}\label{subsec:sampling-distribution}

\begin{figure}[h]
    \centering
    \includegraphics[width=0.99\linewidth]{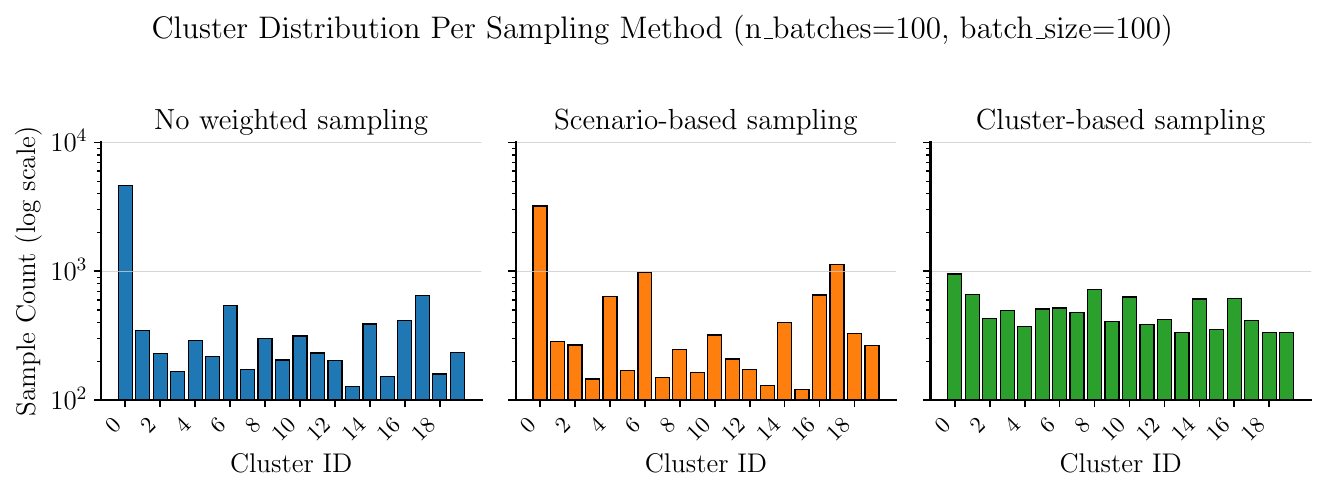}
    \caption{
        Sample counts per cluster after applying different training-time sampling strategies.
        Cluster-based sampling produces the most balanced distributions over the clusters.
    }
    \label{fig:cluster-sampling-comparison}
\end{figure}

\Figref{fig:cluster-sampling-comparison} compares the number of trajectories drawn per cluster across three settings:
no weighted sampling, scenario-based sampling, and cluster-based sampling.
Without any weighting, the samples are heavily skewed toward cluster 0 (stationary trajectories).
Scenario-based sampling slightly reduces the stationary trajectory frequency,
but enhances other frequent clusters (e.g., cluster 4 and 6).
In contrast, cluster-based sampling substantially flattens the cluster distribution and increases coverage of rare motion patterns.

\subsection{Effect of Injection Step on Guided Trajectories}\label{subsec:injection-step}

\begin{figure*}[t]
    \centering
    \begin{subfigure}[t]{0.32\linewidth}
        \centering
        \includegraphics[width=\linewidth]{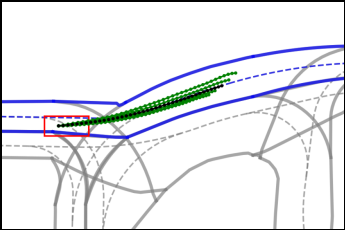}
        \caption{Inject at \(\mathcal{T}_{g,1}=\{\tfrac{1}{4}\}\).}
    \end{subfigure}
    \hfill
    \begin{subfigure}[t]{0.32\linewidth}
        \centering
        \includegraphics[width=\linewidth]{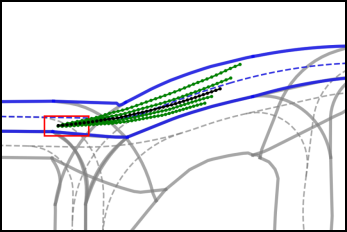}
        \caption{Inject at \(\mathcal{T}_{g,2}=\{\tfrac{1}{2}\}\).}
    \end{subfigure}
    \hfill
    \begin{subfigure}[t]{0.32\linewidth}
        \centering
        \includegraphics[width=\linewidth]{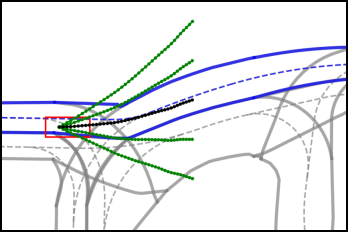}
        \caption{Inject at \(\mathcal{T}_{g,3}=\{\tfrac{3}{4}\}\).}
    \end{subfigure}
    \caption{
        Effect of injecting moderated lateral offsets at different flow steps.
        Black trajectories are with 0 offset,
        green trajectories are with offsets \([-0.5, -0.25, 0.25, 0.5]\).
    }
    \label{fig:moderated-steps}
\end{figure*}

We inject guidance only at a single flow time step (\(|\mathcal{T}_g| = 1\)),
allowing the model to reconcile the perturbation through its learned velocity field at later time steps.
\Figref{fig:moderated-steps} compares injecting \(\delta_{\text{lat}}\) at early, mid, and late flow steps---specifically with
\(\mathcal{T}_{g,1}=\{\tfrac{1}{4}\},\mathcal{T}_{g,2}=\{\tfrac{1}{2}\},\mathcal{T}_{g,3}=\{\tfrac{3}{4}\}\).
Early injections enjoy more time for the model to reconcile context,
whereas very late injections behave similarly to post-hoc shifts and often violate lane-boundary constraints.
Empirically, injecting at \(\mathcal{T}_{g,2}=\{\tfrac{1}{2}\}\) offers the best trade-off between diversity and feasibility.

\subsection{Design explanation}\label{subsec:design-choices}

\paragraph{Ego state injection point.} As highlighted in Diffusion Planner \citep{zheng2025diffusionplanner},
injecting the current ego state at the planner head instead
of as another token to the scene encoder significantly improves performance. We follow this insight: we append it alongside the
noise so the DiT decoder can directly modulate the trajectory with ego-specific cues. This keeps
the encoder focused on exteroceptive context without learning any shortcut trajectory hinted by the ego state.

\paragraph{Post-processing.}  We apply smoothing as we sometimes observe jerky raw trajectories (see \Figref{fig:compare-lane-change})
from FlowDrive. It does not affect the closed-loop performance but affects how the trajectory is scored regarding comfort.
The speed limit is enforced by projecting successive trajectory points along the path and pulling points forward/backward. This
is necessary because human demonstrations often violate speed limits, and imitation learning reproduces these violations (see \Figref{fig:compare-overtake}).

\paragraph{Trajectory lateral divergency vs. injection magnitude and step.}
We quantify the diversity induced by moderated guidance (\Secref{subsec:moderated-guidance}) by measuring the lateral distance
between final positions for symmetric offsets (e.g., $\pm \delta_{\text{lat}}$). As summarized in Table~\ref{tab:lateral-divergency},
divergency increases monotonically with both offset magnitude and later injection step $\mathcal{T}_g$. Adding guidance at early steps
helps more with feasibility (more steps to adapt to the scene), while adding guidance at later steps helps more with diversity
(less remaining steps). A natural extension is to \emph{learn} the injection schedule from scene tokens (lane width, traffic, route),
so the planner stays conservative in routine lane-following while allocating stronger or earlier guidance for overtaking or avoidance.

\begin{table}[t]
    \centering
    \caption{Lateral divergency (m) (lateral distance
between the final positions of the generated two trajectories vs. lateral offset magnitude $[-\delta_{\text{lat}}, \delta_{\text{lat}}]$ and
injection step $\mathcal{T}_g$ (scenario as in \Figref{fig:moderated-steps}).}
    \label{tab:lateral-divergency}
    \footnotesize
    \begin{tabular}{lccc}
        \toprule
        \multirow{2}{*}{Lateral offset magnitudes} & \multicolumn{3}{c}{$\mathcal{T}_g$} \\
        \cmidrule(lr){2-4}
        & $\{\tfrac{1}{4}\}$ & $\{\tfrac{1}{2}\}$ & $\{\tfrac{3}{4}\}$ \\
        \midrule
        $[-0.25,\,0.25]$ & 0.53 & 1.10 & 7.30 \\
        $[-0.5,\,0.5]$   & 1.20 & 2.50 & 15.60 \\
        $[-1,\,1]$       & 2.30 & 4.90 & 29.00 \\
        \bottomrule
    \end{tabular}
\end{table}

\subsection{Ablation Studies}\label{subsec:ablation-studies}

\begin{figure}[t]
    \centering
    \begin{minipage}[c]{0.48\textwidth}
        \centering
        \includegraphics[width=\linewidth]{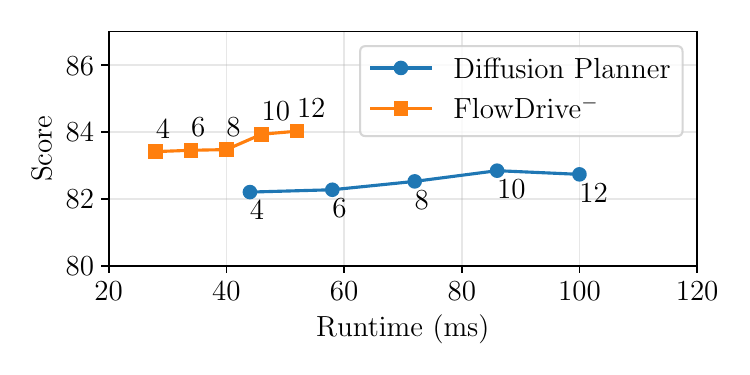}
    \end{minipage}\hfill
    \begin{minipage}[c]{0.48\textwidth}
        \caption{Runtime vs. score on nuPlan Test14 (R) for FlowDrive$^-$ and Diffusion Planner.}
        \label{fig:runtime-vs-score}
    \end{minipage}
\end{figure}

\paragraph{Ablation on Data Balancing Methods.}
We compare the three sampling strategies on nuPlan Val14 (R).
As shown in Table~\ref{tab:ablation-balancing}, scenario-label imbalance is not the main bottleneck:
reweighting by scenario type performs worse than no weighting (FlowDrive$^-$),
while balancing by trajectory pattern (FlowDrive) yields the largest improvement.
We present more detailed scores on different scenarios on Val14 (R)
comparing FlowDrive$^-$ and FlowDrive in Table~\ref{tab:scenario-scores}.

\begin{table*}[t]
    \centering
    \caption{Ablation studies on nuPlan Val14 (R). Best is \textbf{bold}.}
    \vspace{2pt}
    \begin{subtable}[t]{0.4\textwidth}
        \centering
        \caption{Data balancing}
        \footnotesize
        \begin{tabular}{lc}
                \toprule
            Method & Score \\
            \midrule
            No weighted sampling & 81.91 \\
            Scenario-based sampling & 80.08 \\
            Cluster-based sampling & \best{85.37} \\
            \bottomrule
        \end{tabular}
        \label{tab:ablation-balancing}
    \end{subtable}\hfill
    \begin{subtable}[t]{0.6\textwidth}
        \centering
        \caption{Training and inference flow steps}
        \footnotesize
        \begin{tabular}{lccc}
                \toprule
            Training & \multicolumn{3}{c}{Inference flow steps} \\
            flow steps & 6 & 8 & 10 \\
            \midrule
            3000 & 84.84 & 84.76 & 84.81 \\
            2000 & 85.35 & \best{85.37} & 85.28 \\
            1000 & 85.12 & 85.15 & 85.09 \\
            \bottomrule
        \end{tabular}
        \label{tab:ablation-steps}
    \end{subtable}
\end{table*}

\paragraph{Ablation on Flow Training and Inference Steps.}
We use an Euler ODE solver to discretize the flow time.
We train three FlowDrive models with varying training flow steps (1k/2k/3k), all with cluster-based sampling,
and evaluate different numbers of inference flow steps.
As summarized on the left of Table~\ref{tab:ablation-steps},
training with 2k steps and sampling with 8 steps achieves the best score on Val14 (R),
which becomes our default setting throughout this work.

Beyond accuracy, we compare inference-time efficiency between FlowDrive$^-$
    (for a fair comparison without weighted sampling)
    and Diffusion Planner on the nuPlan Test14 (R) split, as shown in \Figref{fig:runtime-vs-score}.
    Across 4--12 steps, both models exhibit consistent performance,
    but FlowDrive$^-$ is typically about twice as fast at the same number of steps due to its first-order solver,
    whereas Diffusion Planner employs a second-order solver that requires two model evaluations per step.
    For instance, at 8 steps FlowDrive$^-$ reaches a score of 83.47 with a runtime of about 40 ms,
    while Diffusion Planner attains 82.90 with about 73 ms.

\paragraph{Ablation on Post-processing Components.}
We study two orthogonal post-processing dimensions on nuPlan Test14 (R):
    (i) candidate generation strategy
    (random sampling using 30 different initial noises vs. our moderated guidance with 30 structured lateral offsets injected mid-flow),
    and (ii) the trajectory scoring function
    (a simple scorer that only retains the progress metric weight vs. the full PDM scorer with additional comfort and time-to-collision terms).
    The results are summarized in Table~\ref{tab:postproc-ablation}.

\begin{table}[t]
    \centering
    \footnotesize
    \caption{Ablation of post-processing components on nuPlan Test14 (R).}
    \vspace{2pt}
    \begin{tabular}{lcc}
        \toprule
        & Random sampling & Moderated guidance \\
        \midrule
        Simple scorer & 90.37 & 92.62 \\
        PDM scorer & 92.37 & 93.96 \\
        \bottomrule
    \end{tabular}
    \label{tab:postproc-ablation}
\end{table}

We observe that moderated guidance yields larger gains than naive random sampling,
    indicating that structured, in-the-loop lateral exploration produces higher-quality, scene-consistent diverse candidates
    than merely increasing the sampled trajectories.
    Furthermore, the improvement from guidance is larger than the improvement from upgrading the scorer,
    showing moderated guidance plays a bigger role in increasing the final performance.

In summary, moderated guidance is a more effective leverage point for performance than naive candidate count increase,
    and its benefits stack with a stronger scoring function, motivating future exploration of learned critics or score functions.

\paragraph{Ablation on design choices.} We ablate several architectural and training choices on the base model FlowDrive and present the results in Table~\ref{tab:design-ablation}.
\begin{itemize}
    \item \textbf{Feature addition vs. concatenation:} Simply adding traffic light, speed limit, and route embeddings in the embedding space
     produces higher scores compared to concatenation followed by a linear projection layer.
    \item \textbf{4s vs. 8s planning horizon:} Increasing the planning horizon to 8 seconds slightly decreases the score.
    \item \textbf{Position vs. velocity action:} Using velocity as the action space instead of position leads to a significant drop, confirming direct
    position prediction is more effective to capture positional interaction between ego vehicle and other lanes, objects.
    \item \textbf{W/o EMA model:} Removing the exponential moving average (EMA) model degrades stability and final performance.
    \item \textbf{W/o pooled token in adaLN-Zero:} Not pooling context tokens for adaptive LayerNorm modulation and only include the
    scene conditional tokens in the Cross-Attention module reduces the model's performance.
    \item \textbf{Uniform vs. lognorm time sampling:} \citet{lee2024improvingtrainingrectifiedflows} proposed a lognorm scale time sampling instead of uniform flow time step
    sampling to improve training stability and let the flow model focus on steps where the velocity field is hard to predict. However,
    this does not help on the final performance of the model.
\end{itemize}

\begin{table}[ht]
    \centering
    \caption{Ablations on nuPlan Val14 (R). Top: design choices. Bottom: number of candidates (left) and number of clusters (right) for cluster-based sampling.}
    \label{tab:ablations}
    \begin{subtable}[t]{1.0\linewidth}
        \centering
        \caption{Design choices}
        \label{tab:design-ablation}
        \footnotesize
        \begin{tabular}{l|c}
            \toprule
            Planner & Score \\
            \midrule
            FlowDrive (base) & 85.37 \\
            vs. feature concat in lane encoder & 83.18 \\
            vs. 8s planning horizon & 84.78 \\
            vs. velocity action & 81.12 \\
            vs. w/o EMA model & 82.76 \\
            vs. w/o pooled token in adaLN-Zero & 82.14 \\
            vs. lognorm sampling & 84.60 \\
            \bottomrule
        \end{tabular}
    \end{subtable}

    \vspace{6pt}
    \begin{subtable}[t]{0.48\linewidth}
        \centering
        \caption{Number of candidates}
        \label{tab:candidates-ablation}
        \footnotesize
        \begin{tabular}{l|c}
            \toprule
            Number of candidates & Score \\
            \midrule
            10 & 0.9226 \\
            20 & 0.9292 \\
            30 & 0.9296 \\
            40 & 0.9258 \\
            \bottomrule
        \end{tabular}
    \end{subtable}\hfill
    \begin{subtable}[t]{0.48\linewidth}
        \centering
        \caption{Number of clusters}
        \label{tab:clusters-ablation}
        \footnotesize
        \begin{tabular}{l|c}
            	\toprule
            Number of clusters ($K$)& Score \\
            \midrule
            5 & 83.45 \\
            10 & 85.08 \\
            20 & 85.37 \\
            30 & 85.38 \\
            \bottomrule
        \end{tabular}
    \end{subtable}
\end{table}

\paragraph{Number of candidates.} We ablate the number of candidates generated by FlowDrive* with different
uniformly distributed lateral offsets between \([-1, 1]\). The results are shown in Table~\ref{tab:candidates-ablation}.

\paragraph{Number of clusters for cluster-based sampling.} On Val14 (R), increasing the number of trajectory-pattern clusters
improves performance of FlowDrive up to about 20 clusters (Table~\ref{tab:clusters-ablation}). Using 5 clusters underfits the motion
modes, 10 clusters recover most of the gains, and 20 clusters essentially saturate. So we adopt $K{=}20$ as a good balance.

\subsection{More Quantitative Results}\label{subsec:more-quantitative-results}

\paragraph{Scores of different scenarios.} Scenario-level closed-loop nuPlan Val14 (R) scores
comparing FlowDrive$^-$ and FlowDrive are presented in Table~\ref{tab:scenario-scores}.
We observe that FlowDrive with cluster-based sampling improves
most motion-centric and turning scenarios significantly without degrading performance on other common scenarios.

\begin{table}[ht]
    \centering
    \caption{Per-scenario nuPlan Val14 (R) scores. Best per row in \best{bold}.}
    \label{tab:scenario-scores}
    \footnotesize
    \begin{tabular}{lcc}
    \toprule
    Scenario (count) & FlowDrive$^-$  & FlowDrive \\
    \midrule
    all (1118) & 81.91 & \best{85.37} \\
    behind\_long\_vehicle (14) & \best{98.54} & 97.98 \\
    changing\_lane (70) & \best{87.16} & 86.12 \\
    following\_lane\_with\_lead (15) & \best{86.32} & 81.24 \\
    high\_lateral\_acceleration (96) & 80.74 & \best{87.17} \\
    high\_magnitude\_speed (99) & 87.72 & \best{90.81} \\
    low\_magnitude\_speed (100) & 78.61 & \best{84.79} \\
    near\_multiple\_vehicles (85) & 78.85 & \best{82.84} \\
    starting\_left\_turn (100) & 67.78 & \best{76.29} \\
    starting\_right\_turn (98) & 76.47 & \best{81.40} \\
    traffic\_light\_intersection\_traversal (98) & 81.04 & \best{81.75} \\
    stationary\_in\_traffic (98) & 91.37 & \best{93.95} \\
    stopping\_with\_lead (93) & \best{96.87} & 95.98 \\
    traversing\_pickup\_dropoff (99) & 77.19 & \best{80.30} \\
    waiting\_for\_pedestrian\_to\_cross (53) & 75.14 & \best{80.18} \\
    \bottomrule
    \end{tabular}
\end{table}

\begin{figure*}[t]
    \centering
    \includegraphics[width=1.0\linewidth]{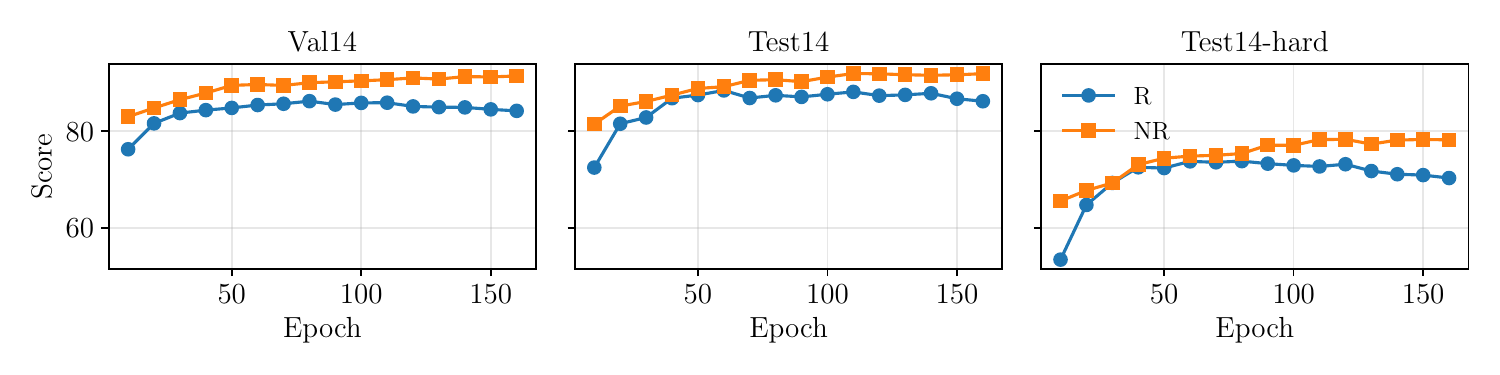}
    \caption{Training epoch trends for FlowDrive on Val14, Test14 and Test14-hard (R vs. NR).}
    \label{fig:epoch-trends}
\end{figure*}

\paragraph{Scores on NR and R benchmarks of different epochs.} During the experiments, we find that the NR-scores of
FlowDrive on nuPlan improve steadily and saturate around epoch 120; the R-scores peak slightly earlier (\(\sim\)60--80) and
then decrease if training continues (\Figref{fig:epoch-trends}). This indicates that during training, the model gradually memorizes the
pattern of how the surrounding vehicles behave and thus improves NR performance steadily, but eventually overfits to the training data
and loses generalization ability in the R setting. We therefore select the checkpoint at epoch 110 rather than the final epoch (400).
All reported results in this paper for FlowDrive$^-$ and FlowDrive use the epoch-110 checkpoint.

\subsection{Qualitative results}\label{subsec:qualitative-challenging}
\begin{figure*}[t]
    \centering
    \includegraphics[width=1.0\linewidth]{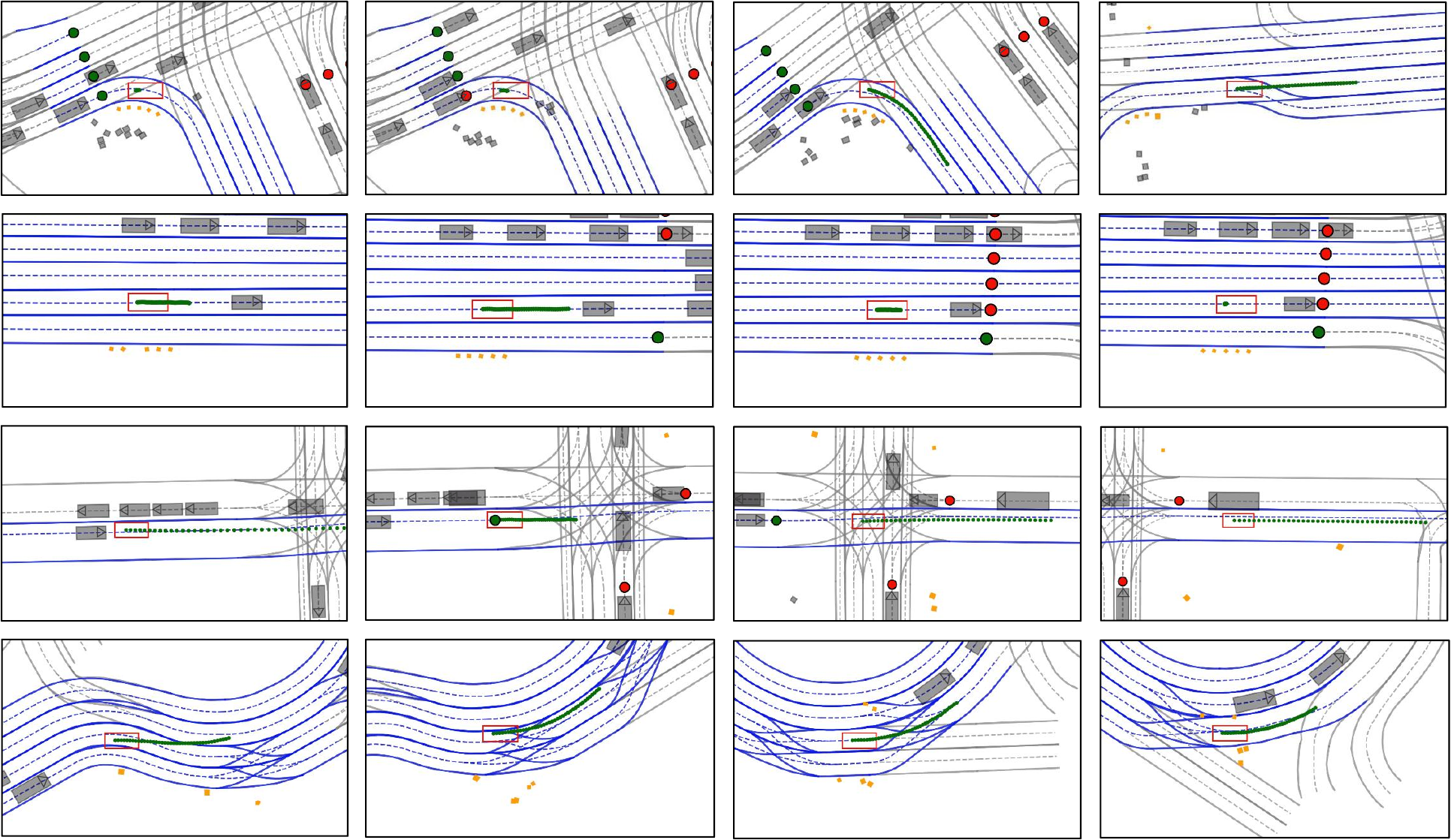}
    \caption{Qualitative examples of FlowDrive* on nuPlan scenarios (R-mode).
    Each row represents a different scenario at 0, 3.3, 6.6, and 10 seconds intervals. The red box is the ego vehicle.
    Green indicates the planned trajectory. The gray boxes are other dynamic agents with an overlapping
    triangle indicating their driving direction. Red and green circles represent
    traffic lights (red and green respectively). First row: a right turn at an intersection with pedestrians crossing.
    Second row: car-following behavior with a final stop. Third row: slowing down for crossing vehicles.
    Fourth row: navigation through narrow pickup/dropoff area while overtaking slow vehicles on the left lane.}
    \label{fig:nuplan-qua}
\end{figure*}

\begin{figure*}[t]
    \centering
    \includegraphics[width=1.0\linewidth]{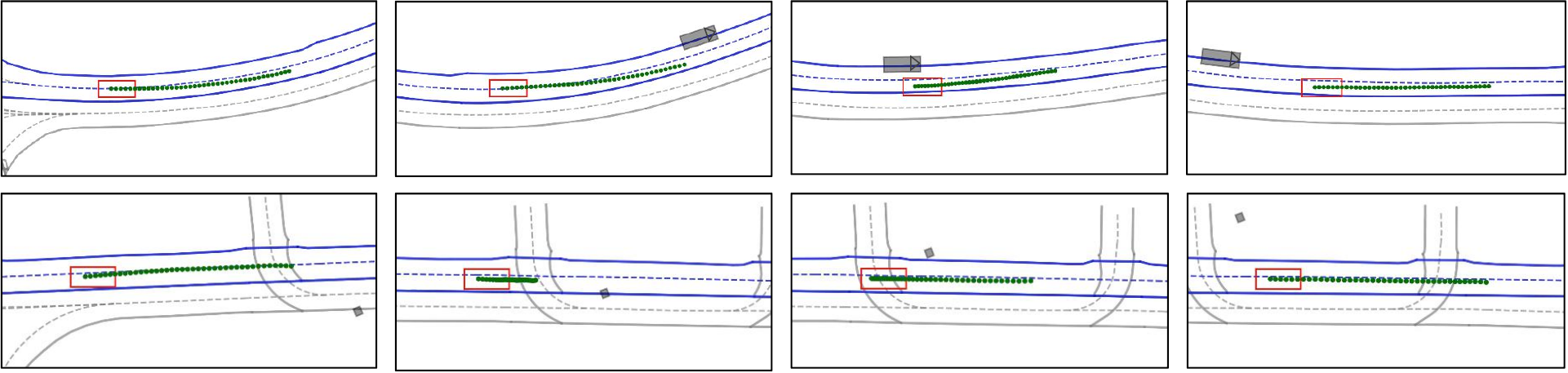}
    \caption{Qualitative examples of FlowDrive* on interPlan scenarios.
    First row: nudging around a parked vehicle at 4, 8, 12, and 16 seconds.
    Second row: emergency stop for a suddenly crossing pedestrian at 2, 4, 6, and 8 seconds.}
    \label{fig:interplan-qua}
\end{figure*}

\Figref{fig:nuplan-qua} and \Figref{fig:interplan-qua} show qualitative examples from nuPlan and interPlan scenarios.
FlowDrive* demonstrates robust performance across diverse driving situations, including waiting for pedestrians,
intersection navigation, car-following, and complex maneuvers like nudging around parked vehicles and emergency stopping.

\begin{figure*}[t]
    \centering
    \begin{subfigure}[t]{1.0\linewidth}
        \centering
        \includegraphics[width=\linewidth]{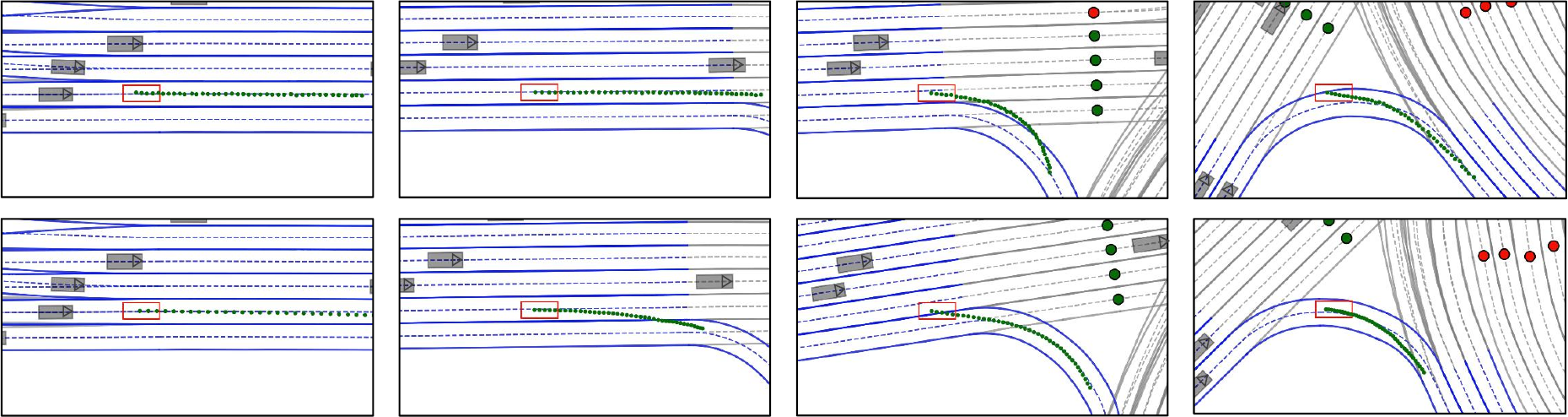}
        \caption{Lane change scenario from interPlan at 0, 3.3, 6.6, and 10 seconds. First row: FlowDrive changes lanes too late
        to follow the route, and finally drives out of the lane.
        Second row: FlowDrive* successfully changes lanes to follow the route.}
        \label{fig:compare-lane-change}
    \end{subfigure}

    \vspace{10pt}

    \begin{subfigure}[t]{1.0\linewidth}
        \centering
        \includegraphics[width=\linewidth]{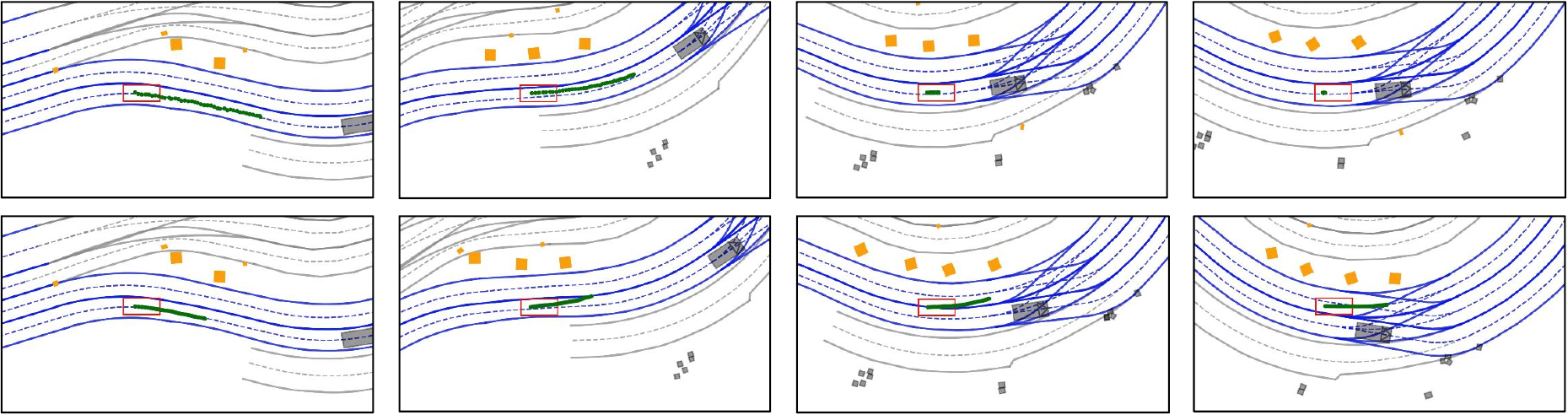}
        \caption{Overtaking scenario from nuPlan at 0, 5, 10, 15 seconds. First row: FlowDrive fails to overtake the
        slow leading vehicle in a pickup/dropoff area. It also starts with a higher velocity than the speed limit at 0 seconds.
        Second row: FlowDrive* successfully overtakes the slow leading vehicle while maintaining the speed limit at all times.}
        \label{fig:compare-overtake}
    \end{subfigure}

    \caption{Comparison between FlowDrive and FlowDrive* in challenging scenarios.}
    \label{fig:flowdrive-comparison}
\end{figure*}

\Figref{fig:flowdrive-comparison} compares FlowDrive and FlowDrive* on challenging scenarios. \Figref{fig:compare-lane-change}
shows a lane change scenario where FlowDrive changes lane too late to follow the route and eventually drives
out of the lane, while FlowDrive* successfully executes the lane change maneuver. \Figref{fig:compare-overtake} demonstrates
an overtaking scenario where FlowDrive fails to overtake a slow leading vehicle in a pickup/dropoff area and
also violates speed limits, whereas FlowDrive* successfully overtakes while maintaining proper speed limits
throughout the maneuver.

\begin{figure*}[t]
    \centering
    \includegraphics[width=1.0\linewidth]{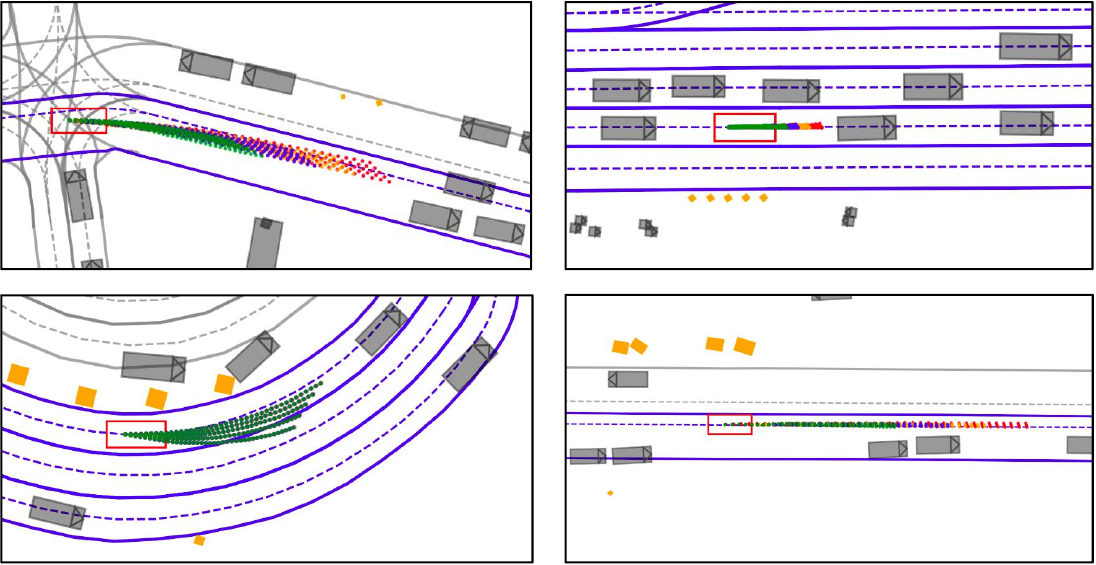}
    \caption{Jointly moderated guidance with lateral (same color) and longitudinal offsets (different color).
    Top left: overtake parked vehicles with different speed and lateral offsets.
    Top right: car-following with varied longitudinal offsets.
    Bottom left: longitudinal offsets do not change final speeds, while lateral offsets induce overtaking.
    Bottom right: on narrow roads, all lateral offsets stay collision-free relative to lane boundaries and parked vehicles.}
    \label{fig:guidance}
\end{figure*}

\end{document}